\documentclass[10pt,twocolumn,letterpaper]{article}

\usepackage{iccv}
\usepackage{times}
\usepackage{epsfig}
\usepackage{graphicx}
\usepackage{amsmath}
\usepackage{amssymb}

\usepackage{booktabs}
\usepackage{inconsolata}
\usepackage{multirow}
\usepackage{pifont}

\newcommand{\cmark}{\ding{51}}%
\newcommand{\xmark}{\ding{55}}%

\usepackage[pagebackref=true,breaklinks=true,letterpaper=true,colorlinks,bookmarks=false]{hyperref}

\iccvfinalcopy 

\def\iccvPaperID{1105} 

\ificcvfinal\fi

\begin{document}

\title{End-to-End Active Speaker Detection }

\author{
    Juan León Alcázar$^{1}$, Moritz Cordes$^{2}$, Chen Zhao$^{1}$ \& Bernard Ghanem$^{1}$\\
    \small $^{1}$ King Abdullah University of Science and Technology (KAUST), $^{2}$Leuphana University of Lüneburg\\
    {\tt\small jc.leon@uniandes.edu.co, moritz.cordes@stud.leuphana.de, chen.zhao@kaust.edu.sa, bernard.ghanem@kaust.edu.sa}
}

\maketitle
\ificcvfinal\thispagestyle{empty}\fi

\begin{abstract}
    Recent advances in the Active Speaker Detection (ASD) problem build upon a two-stage process: feature extraction and spatio-temporal context aggregation. In this paper, we propose an end-to-end ASD workflow where 
    feature learning and contextual predictions are  jointly learned. Our end-to-end trainable     network simultaneously learns multi-modal embeddings and aggregates spatio-temporal context. This results in more suitable feature representations and improved performance in the ASD task. We also introduce interleaved graph neural network (iGNN) blocks, which split the message passing according to the main sources of context in the ASD problem.
    Experiments show that the aggregated features from the iGNN blocks are more suitable for ASD, resulting in state-of-the art performance. 
    Finally, we design a weakly-supervised strategy, which demonstrates that the ASD problem can also be approached by utilizing audiovisual data but relying exclusively on audio annotations. We achieve this by  modelling the direct relationship between the audio signal and the possible sound sources (speakers), as well as introducing a contrastive loss. All the resources of this project will be made available at: \url{https://github.com/fuankarion/end-to-end-asd}.
\end{abstract}


\section{Introduction}

\label{sec:intro}
 \begin{figure*}[t]
    \begin{center}
        \includegraphics[width=0.99\textwidth]{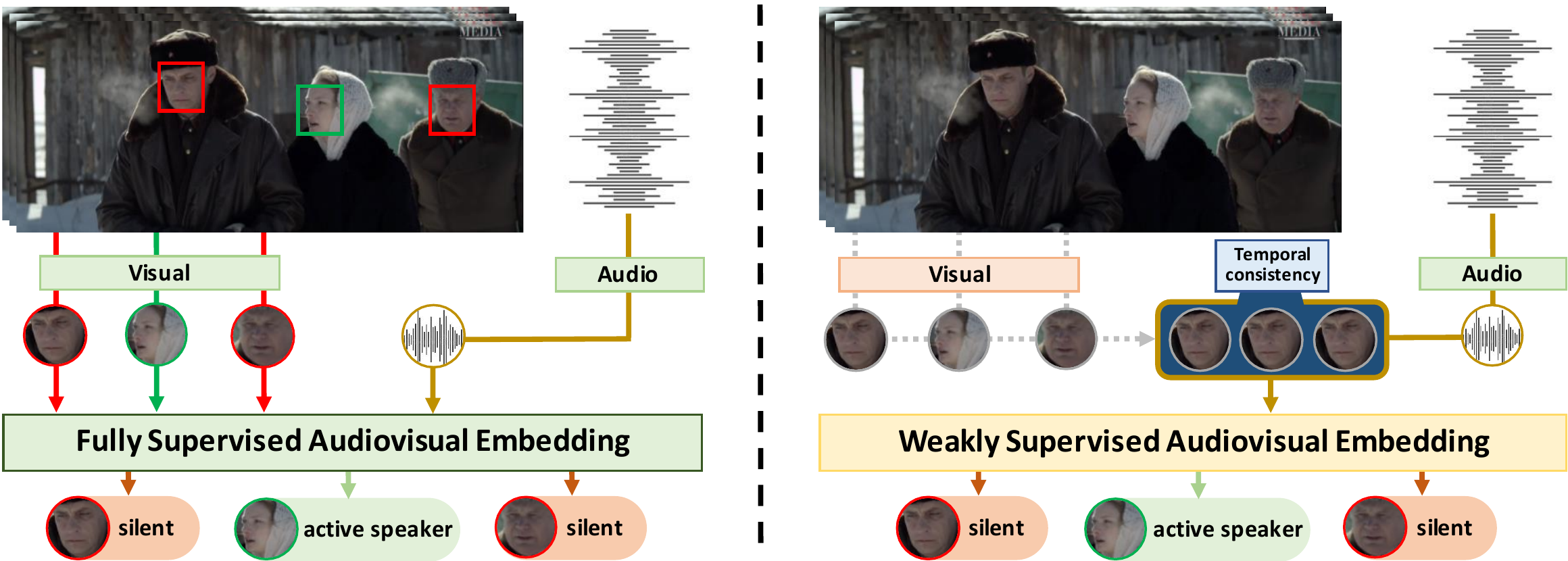}        
    \end{center}
    \caption{
        \textbf{Fully and weakly-supervised audiovisual embeddings.} In the fully supervised scenario (left), we use the face crops as visual data and the Mel-frequency cepstral coefficients as audio data, we rely on visual and audio labels to directly optimize a shared feature embedding. In contrast, in the weakly supervised scenario, we omit the visual labels and optimize using only audio supervision. By modeling the visual-temporal consistency and speech-to-speaker assignments, we are able to optimize a shared embedding that can detect the active speakers without any visual supervision.
    }
    \label{fig:firstfig}
\end{figure*}
In active speaker detection (ASD), the current speaker must be identified from a set of available candidates, which are usually defined by face tracklets assembled from temporally linked face detections~\cite{roth2020ava,chakravarty2016active,leon2021maas}. Initial approaches to the ASD problem focused on the analysis of individual visual tracklets and the associated audio track, aiming to maximize the agreement between the audio signal and the visual patterns \cite{roth2020ava,chung2019naver,zhangmulti}. Such an approach is suitable for scenarios where a single visual track is available. However, in the general (multi-speaker) scenario, this naive correspondence will suffer from false positive detections, leading to incorrect speech-to-speaker assignments.

Current approaches for ASD rely on two-stage models \cite{leon2021maas,kopuklu2021design,tao2021someone}. First, they associate the facial motion patterns and its concurrent audio stream by optimizing a multi-modal encoder \cite{roth2020ava}. This multi-modal encoder serves as a feature extractor for a second stage, in which multimodal embeddings from multiple speakers are fused \cite{alcazar2020active}. 

These two-stage approaches are currently preferred given the technical challenges of end-to-end training with video data. Despite the computational efficiency of these approaches, their two-stage nature precludes them from 
fully leveraging the learning capabilities of modern neural architectures, namely directly optimizing the features for the multi-speaker ASD task. 

In this paper, we present a novel alternative to the traditional two-stage ASD methods, called End-to-end Active Speaker dEtEction (EASEE), which is the first end-to-end pipeline for active speaker detection. Unlike conventional methods, EASEE is able to learn multi-modal features from multiple visual tracklets, while simultaneously modeling their spatio-temporal relations in an end-to-end manner. As a consequence, EASEE feature embeddings are optimized to capture information from multiple speakers and enable effective speech-to-speaker assignments in a fully supervised manner. To generate its final predictions, our end-to-end architecture relies on a spatio-temporal module for context aggregation. We propose an interleaved Graph Neural Network (iGNN) block to model the relationships between speakers in adjacent timestamps. Instead of greedily fusing all available feature representations from multiple timestamps, the iGNN block provides a more principled way of modeling spatial and temporal interactions. iGNN performs two message passing steps: first a spatial message passing that models local interactions between speakers visible at the same timestamp, and then a temporal message passing that effectively aggregates long-term temporal information.

Finally, EASEE's end-to-end nature allows the use of alternative supervision targets. In this paper, we propose a weakly-supervised strategy for ASD, named EASEE-W (shown in Figure~\ref{fig:firstfig}). 
EASEE-W relies exclusively on audio labels, which are easier to obtain, to train the whole architecture. To optimize our network without the visual labels, we model the inherent structure in the ASD task, namely the direct relationship between the audio signal and its possible sound sources, \ie, the speakers.

\noindent \textbf{Contributions.} This paper proposes EASEE, a novel strategy for active speaker detection. Its end-to-end nature enables direct optimization of audio-visual embeddings and leverages novel training strategies, namely weak supervision. Our work brings the following contributions: (1) We devise \textbf{the first end-to-end trainable neural architecture} EASEE for the active speaker problem (Section \ref{subsec:Architecture}), which learns effective feature representations. (2) In EASEE, we propose \textbf{a novel iGNN block} to aggregate spatial and temporal context based on a composition of spatial and temporal message passing. We show this reformulation of the graph structure is key to achieve state-of-the-art results (Section \ref{subsec:sota}). (3) Based on EASEE, we propose \textbf{the first weakly-supervised ASD approach} that enables the use of only audio labels to generate predictions on visual data (Section \ref{subsec:weaksupervision}).

\section{Related Work}

Early approaches to the ASD problem \cite{cutler2000look} attempted to correlate audiovisual patterns using time-delayed neural networks \cite{waibel1989phoneme}. Follow up works \cite{saenko2005visualspeech,everingham2009taking} approached the ASD task by limiting the analysis only to visual patterns. These approaches rely only on visual data given the biases of the single speaker scenario (\ie speech can only be attributed to the single visible speaker). A parallel corpus of work focused on the complementary task of voice activity detection (VAD), which aims at finding speech activities among other acoustic events \cite{tanyer2000voice,chang2006voice}. Similar to visual data, audio-only information was also proven to be useful in single speaker scenarios \cite{ding2019personal}.

The recent interest in deep neural architectures \cite{rumelhart1986learning,lecun1989handwritten,krizhevsky2012imagenet} shifted the focus in the ASD problem from hand-crafted feature design to multi-modal representation learning \cite{ngiam2011multimodal}. As a consequence, ASD has become dominated by CNN-based approaches, which rely on convolutional encoders originally devised for image analysis tasks \cite{roth2020ava}. Recent works \cite{chakravarty2016active,chung2016out} approached the more general multi-speaker scenario, relying on the fusion of multi-modal information from individual speakers. Concurrent works have also focused on audiovisual feature alignment. This resulted in methods that rely on audio as the primary source of supervision \cite{chakravarty2015s}, or focused on the design of multi-modal embeddings \cite{chung2018voxceleb2,chung2016out,nagrani2017voxceleb,tao2017bimodal}.

 \begin{figure*}[t]
    \begin{center}
        \includegraphics[width=0.99\textwidth]{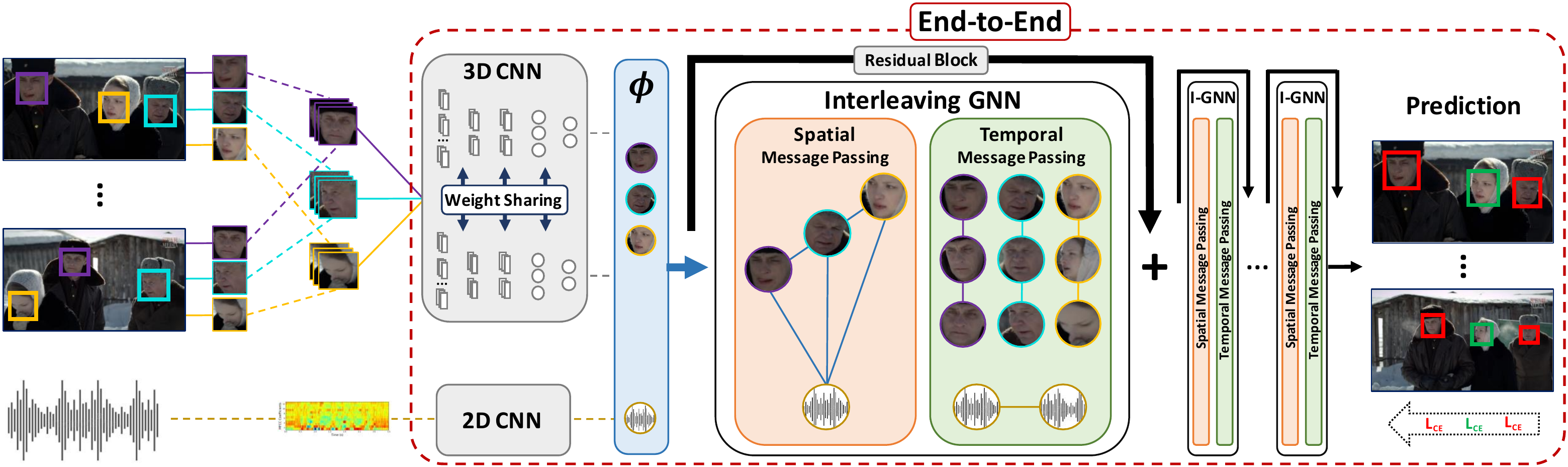}
    \end{center}
    \caption{
        \textbf{Overview of the EASEE architecture.}  We fuse information from multiple visual tracklets, and their associated audio track. We rely on a 3D CNN to encode individual face tracklets, and a 2D CNN to encode the audio stream (Gray Encoders). These embeddings are assembled into an initial multi-modal embedding ($\Phi$) containing audiovisual information from multiple persons in a scene. We map this embedding into a graph structure that performs message passing steps over spatial (light orange) and temporal dimensions (light green). Our layer arrangement favors independent massage passing steps along the temporal and spatial dimensions. 
    }
    \label{fig:overview}
\end{figure*}

The recent availability of large-scale data for the ASD task \cite{roth2020ava} has enabled the use of state-of-the-art deep convolutional encoders \cite{he2016deep,hara2018can}. In addition to these deep encoders, current approaches have shifted focus to directly modeling the temporal features over short temporal windows, typically by optimizing a Siamese Network with modality specific streams. The work of Chung \etal \cite{chung2019naver} explored the use of a hybrid 3D-2D encoder pretained on VoxCeleb \cite{chung2018voxceleb2} to analyze these temporal windows, while Zhang \etal \cite{zhangmulti} focused on improving the feature representation by using a contrastive loss \cite{hadsell2006dimensionality} between the modalities. 

To complement this short-term analysis, many methods \cite{kopuklu2021design,leon2021maas,tao2021someone} have aimed to incorporate contextual information from overlapping visual tracklets. The work of Alcazar \etal \cite{alcazar2020active} introduced a data structure to represent an active speaker scene, and the features in this structure are improved by using self-attention\cite{wang2018non,vaswani2017attention} and recurrent networks \cite{hochreiter1997long}. 

Current state-of-the-art techniques incorporate contextual representation and rely on deep 3D encoders for the initial feature encoding and recurrent networks or self-attention to analyze the scene's contextual information \cite{leon2021maas,kopuklu2021design,tao2021someone,zhang2021unicon}. We depart from this standard approach and devise a strategy to train end-to-end networks that simultaneously optimize features from a shared multi-modal encoder. This enables the direct optimization of temporal and spatial features for the ASD problem in a multi-speaker setup.  

\subsection{Graph Convolutional Networks}
The current interest in non-Euclidean data \cite{Gkioxari2019Mesh,cv_action_jain2016structural,cv_inv_scene_johnson2018image,li2019sgas,cv_scene_li2018factorizable,g_tad} has focused the attention of the research community on Graph Convolutional Networks (GCNs) as an efficient variant of CNNs  \cite{kipf2016semi,wu2019simplifying}. GCNs have achieved state-of-the-art results in zero-shot recognition \cite{wang2018zero,kampffmeyer2019rethinking}, 3D understanding \cite{Gkioxari2019Mesh,li2019sgas,xie2019clouds}, and action recognition in video \cite{cv_action_jain2016structural,g_tad,cv_action_yan2018spatial} by harnessing the flexibility of graphs representations. Recently, GCNs have been widely used in the field of action recognition, focusing on skeleton-based approaches that rely only on visual data \cite{cai2021jolo,duhme2021fusion}. For applications in audiovisual contexts, GCNs have been utilized to study inter-correlations in videos for automatic recognition of emotions in conversations \cite{nie2020c,ren2021lr}. In the ASD domain, Alcazar \etal \cite{leon2021maas} introduced the use of GCNs, developing a two-stage approach where a GCN network would module interactions between audio and video across multiple frames. We present an alternative to this approach where we focus on the end-to-end modeling, and perform independent steps of message passing along the spatial and temporal dimensions.

\section{End-to-End Active Speaker Detection}
Our approach relies on the initial generation of independent audio and visual embeddings at specific timestamps. These embeddings are fused and jointly optimized by means of a graph convolutional network \cite{kipf2016semi}. To this end, we devise a neural architecture with three main components: (i) audio Encoder, (ii) visual Encoder, and a (iii) spatio-temporal Module. The visual encoder ($f_{v}$) performs multiple forward passes (one for each available tracklet), and the audio encoder ($f_{a}$) performs a single forward pass on the shared audio clip. These features are arranged according to their temporal order and (potential) spatial overlap, creating an intermediate feature embedding ($\Phi$) that enables spatio-temporal reasoning. Unlike other methods, we construct $\Phi$ such that it can be optimized end-to-end. Thus $\Phi$ captures multi-modal and multi-speaker information, enables information flow across modalities, and ultimately improves network predictions. Figure \ref{fig:overview} contains an overview of our proposed approach.

 \begin{figure*}[t]
    \begin{center}
        \includegraphics[width=0.99\textwidth]{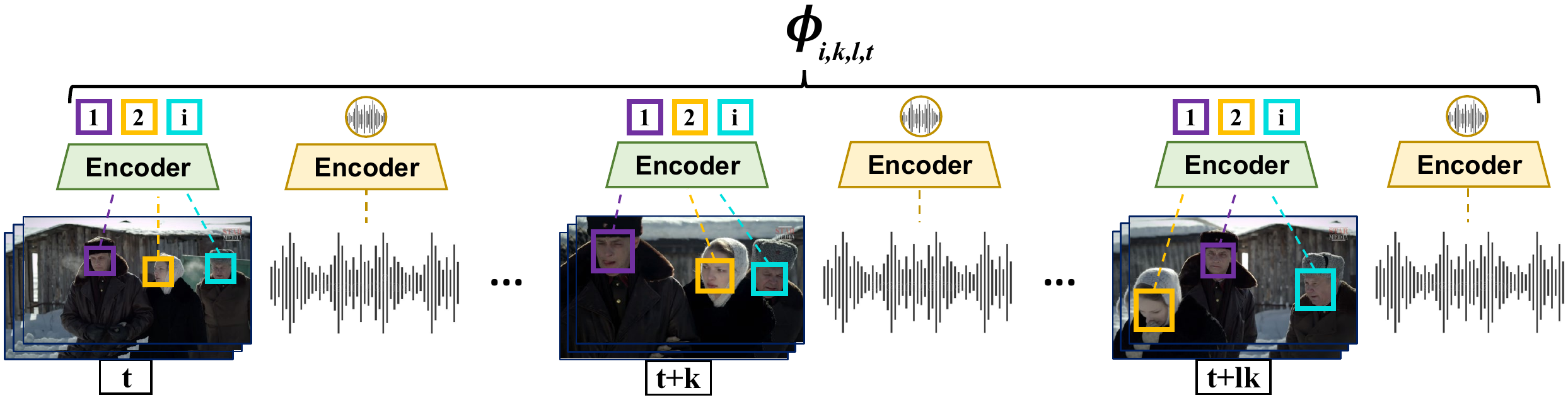}
    \end{center}
    \caption{ 
        \textbf{EASEE Sub-Sampling.} For every temporal endpoint, we sample $i$ face tracklets and the corresponding audio signal. This sampling is repeated over $l$ consecutive temporal endpoints separated by stride $k$. The $i+1$ feature embeddings obtained at each timestamp are forwarded through the audio (yellow) and visual (light green) encoders fused into the spatio-temporal embedding $\Phi_{i,k,l,t}$.
    }
    \label{fig:sampling}
\end{figure*}

\subsection{EASEE Network Architecture} 
\label{subsec:Architecture}

The main goal of EASEE is to aggregate related temporal and spatial information from different modalities over a video segment. To enable efficient end-to-end computation, we do not densely sample  all the available tracklets in a temporal window, but rather define a strategy to sub-sample audiovisual segments inside a video. We define a set of temporal endpoints where the original video data (visual and audio) is densely sampled. At every temporal endpoint, we collect visual information from the available face tracklets and sample the associated audio signal  (See Figure \ref{fig:sampling}). To further limit the memory usage, we define a fixed number of tracklets ($i$) to sample at every endpoint. Since the visual stream might contain an arbitrary number of tracklets, we follow \cite{alcazar2020active} at training time and sample $i$ tracklets with replacement. Hence, from every temporal endpoint, we create $i+1$ feature embeddings associated with it ($i$ visual embeddings from $f_{v}$ and the audio embedding from $f_{a}$). 

We create temporal endpoints over a video segment following a simple strategy, we select a timestamp $t$ and create $l$ temporal endpoints over the video at a fixed stride of $k$ frames. The location of every endpoint is then given by $L = \{t, t+k, ..., t+lk\}$. This reduces the total number of samples from the video data by a factor of $k$ and allows us to sample longer sections of video for training and inference. 

\paragraph{Spatio-Temporal Embedding.} We build the embedding $\Phi$ over the endpoint set $L$. We define the spatio-temporal embedding $e$ at time $t$ for speaker $s$ as $e_{t,s} = \{f_{a}(t), f_{v}(s, t)\}$. Since there may be multiple visible persons at this endpoint (\ie $|s| \geq 1 $), we define the embedding for an endpoint at time $t$ with up to $i$ speakers  as $E_{t,i} = \{e_{t,0}, e_{t,1}, e_{t,2}, ..., e_{t,i}\}$. The full spatio-temporal embedding $\Phi_{i,k,l,t}$ is created by sampling audio and visual features over the endpoint set $L$, thus $\Phi_{i,k,l,t} = \{ E_{t,i}, ..., E_{t+k,i}, ..., E_{t+lk,i}\}$. As $\Phi_{i,k,l,t}$ is assembled from independent forward passes of the $f_{a}$ and $f_{v}$ encoders, we share weights for forward passes in the same modality, thus each forward/backward pass accumulates gradients over the same weights. This shared weight scheme largely simplifies the complexity of the proposed network, and keeps the total number of parameters stable regardless of the values for $l$ and $i$.

Upon computing the initial modality embeddings, we map $\Phi_{i,k,l,t}$ into a spatio-temporal graph representation. Following \cite{leon2021maas}, we map each feature in $\Phi_{i,k,l,t}$ into an individual node, resulting in a total of $(i+1)*l$ nodes. Every feature embedding goes through a linear layer for dimensionality reduction before being assigned to a node. Unlike \cite{leon2021maas}, we are not interested in building a unique graph structure that performs message passing over all the possible relationships in the node set. Instead, we choose to independently model the two types of information flow in the graph, namely spatial information and temporal information.

\subsection{Graph Neural Network Architecture}
In EASEE, the GCN component fuses spatio-temporal information from video segments. This module implements a novel composition pattern where the spatial and temporal information message passing are performed in subsequent layers. We devise a building block (iGNN) where the spatial message passing is performed first, then temporal message passing occurs. After these two forward passes, we fuse the feature representation with the previously estimated feature embedding (residual connection). We define the iGNN block at layer $J$ as: 

\vspace{-0.3cm}
\begin{eqnarray*}
    \Phi^{s} = M^{s} (A^{s}\Phi^{J};\theta^{s}) , \Phi^{t} = M^{t} (A^{t}\Phi^{J};\theta^{t}) \\
    iGNN(\Phi^{J}) = (M^{t} \circ M^{s})(\Phi^{J}) + \Phi^{J} \\
\end{eqnarray*}
\vspace{-0.6cm}

Here, $M^{s}$ is a GCN layer that performs spatial message passing using the spatial adjacency matrix $A^{s}$ over an initial feature embedding ($\Phi^{J}$), thus producing an intermediate representation with aggregated local features ($\Phi^{J+1}$). Afterwards, the GCN layer $M^{t}$ performs a temporal message passing using the temporal adjacency matrix $A^{t}$. $\theta^{s}$ and $\theta^{t}$ are the parameter set of their respective layers. The final output is complemented with a residual connection, thus favoring gradient propagation. 

\begin{figure}[t]
    \begin{center}
        \includegraphics[width=0.48\textwidth]{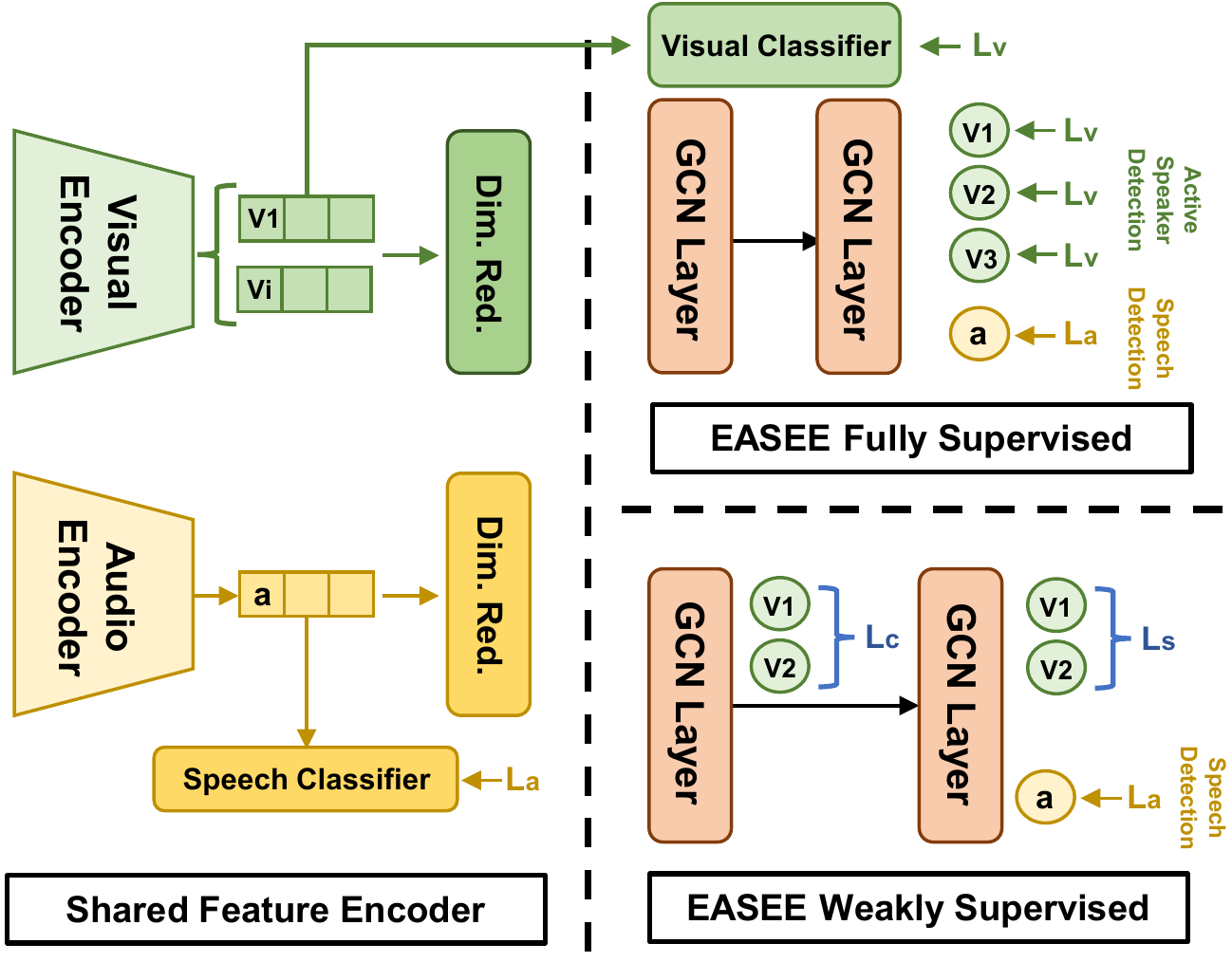}        
    \end{center}
    \caption{
        \textbf{EASEE Weakly Supervised.} We drop all the visual supervision ($\mathcal{L}_{v}$) in EASEE (intermediate supervision included) and enforce positive predictions amongst the video nodes (light green) in the presence of a speech event ($\mathcal{L}_{s}$), along with consistent visual feature representations if the nodes contain the same identity
    }
    \label{fig:weakly}
\end{figure}

In EASEE, the assignment of elements from the embedding $\Phi_{i,k,l,s}$ to graph nodes remains stable throughout the entire GCN structure (\ie we do not perform any pooling). This allows us to create a final prediction for every tracklet and audio clip contained in $\Phi_{i,k,l,t}$ by applying a single linear layer. This arrangement creates two types of nodes: \textit{Audio Nodes}, which generate predictions for the audio embeddings (\ie speech detected or silent scene), and \textit{Video Nodes} which generate predictions for the visual tracklets (\ie active speaker or silent). EASEE's final predictions are made only from the output of visual nodes.

\paragraph{Losses \& Intermediate Supervision.} Audio nodes are supervised in training, but their forward phase output is not suitable for the ASD task. The training loss is defined as: $\mathcal{L} = \mathcal{L}_{a}+ \mathcal{L}_{v}$. where $\mathcal{L}_{a}$ is the loss over all audio nodes and $\mathcal{L}_{v}$ is the loss over all the video nodes. $\mathcal{L}_{a}$ and $\mathcal{L}_{v}$ are implemented as  cross-entropy (CE) losses.  iGNN is also supervised with  CE loss, which is calculated individually at every node in the last layer. 

\subsection{Weakly Supervised Active Speaker Detection} 

\label{sec:weakly}
State-of-the-art methods rely on fully supervised approaches to generate consistent predictions in the ASD problem. Typically, they work in a fully supervised manner in both learning stages, using audiovisual labels to train the initial feature encoder and also to supervise the second stage learning \cite{kopuklu2021design,tao2021someone,alcazar2020active,leon2021maas}. The end-to-end nature of EASEE enables us to approach the active speaker problem from a novel perspective, where the multi-speaker scenario can be analyzed relying on a weak supervision signal, namely only audio labels. In comparison to visual labels, audio ground-truth is less expensive to acquire, as it only establishes the start and end point of a speech event. Meanwhile, labels for visual data must establish the fine-grained association between every temporal interval in the speech event and its visual source. 

Directly training EASEE with audio labels only, would optimize the predictions for the audio nodes (speech events). As outlined before, such predictions are suitable for the voice activity detection task, but the more fine grained ASD task will have poor performance as the visual nodes lack any supervision and yield random outputs. To generate meaningful predictions for the visual nodes while relying only on audio supervision, we reformulate our end-to-end training to enforce information flow between modalities by adding two extra loss functions on the graph structure. This reformulation enables meaningful predictions over the visual data despite the lack of visual ground-truth. We name this version of our approach EASEE-W, a novel architecture that is capable of making active speaker predictions that rely only on weak binary supervision labels from the audio stream. An overview of the key differences between EASEE and EASEE-W is shown in Figure \ref{fig:weakly}.

\paragraph{Local assignment loss.} We design a loss function that models local dependencies in the ASD problem: if there is a speech event, we must attribute the speech to one of the locally associated video nodes. Let $V_{t}$ be the output of video nodes at time $t$ ($|V_{t}| \geq 2$), and $y_{at}$ the ground truth for the audio signal at time $t$:
\begin{eqnarray*}
    L_{s} = y_{at}(y_{at} - \max(V_{t})) + (1-y_{at})\max(V_{t})
\end{eqnarray*}

 \begin{figure}[t]
    \begin{center}
        \includegraphics[width=0.45\textwidth]{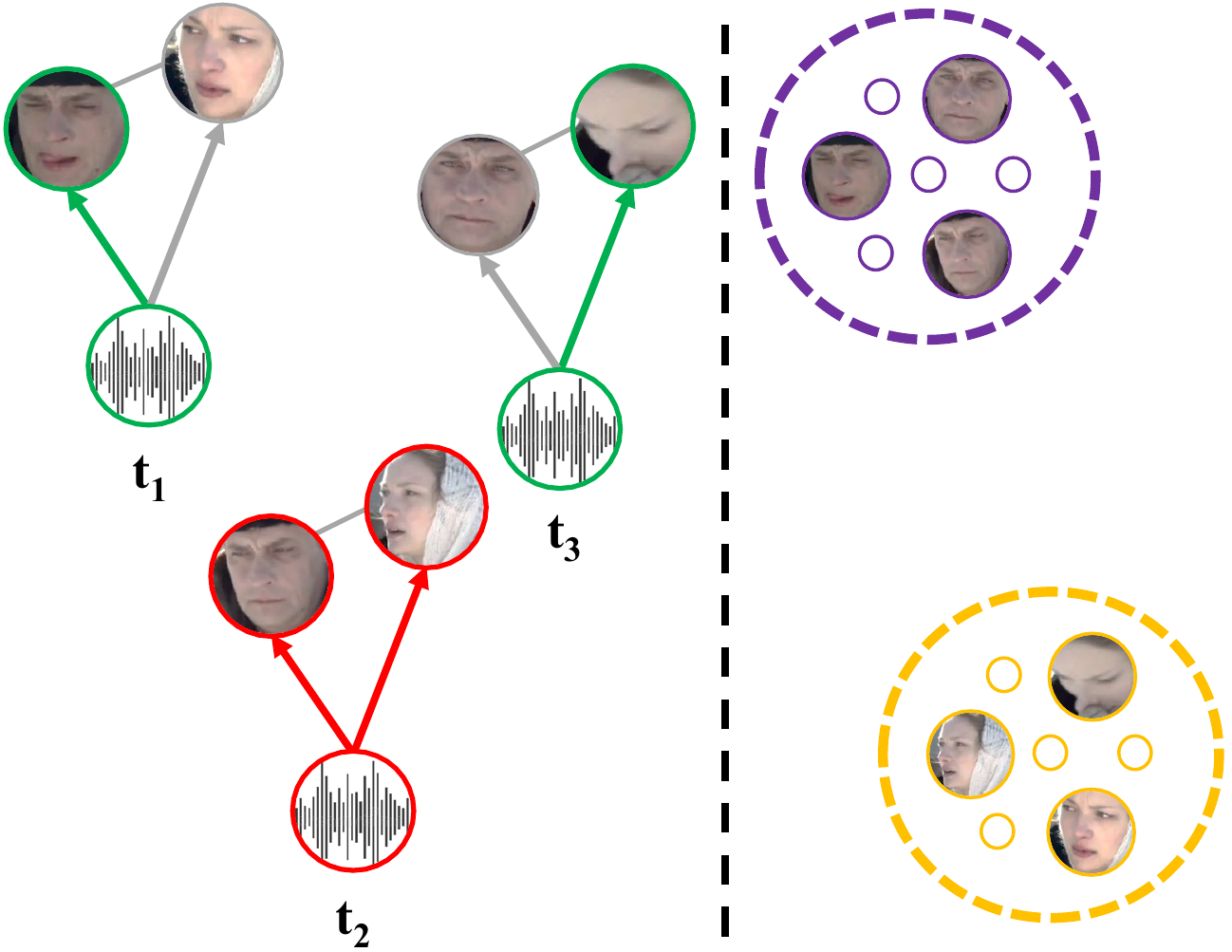}        
    \end{center}
    \caption{
        \textbf{Weakly Supervised Losses.} We enforce an individual speaker assignment if there is a detected speech event (left). Temporal consistency pulls together features for faces of the same person and creates differences for faces of different persons (right).
    }
    \label{fig:VCL}
\end{figure}

The first term $y_{at}(y_{at} - \max(V_{t}))$ will force EASEE-W to generate at least one positive prediction in $V_{t}$ if $y_{at}=1$ (\ie select a speaker if speech is detected). Likewise, the second term $(1-y_{at})\max(V_{t})$ will force EASEE-W to generate only negative predictions in $V_{t}$ in the absence of speech. While this loss forces the network to generate video labels that are locally consistent with the audio supervision, we show that these predictions only improve the performance over a fully random baseline and do not represent an improvement over trivial audiovisual assignments. 

\paragraph{Visual contrastive loss.} We complement $L_{s}$ with a contrastive loss ($L_{c}$) applied over the video data. As shown in Figure \ref{fig:VCL}, the goal of this loss is to enforce feature similarity between video nodes that belong to the same person, and promote feature differences for non-matching identities. Considering that the AVA-ActiveSpeaker dataset \cite{roth2020ava} does not include identity meta-data, we approximate the sampling of different identities by selecting visual data from concurrent tracklets\footnote{Since tracklets include a single face and were manually curated, it is guaranteed that two tracklets that overlap in time belong to different identities. If there is a single person in the scene, we sample additional visual data from another tracklet in a different movie where no speech event is detected.}. To simplify the contrastive learning, we modify the sampling scheme for EASEE-W, and force $i=2$ regardless of the real number of simultaneous tracklets. If there are more than 2 visible persons in the scene, we just sample without replacement.

In practice, we follow \cite{chen2020simple} and apply this loss on the second to last layer of the iGNN block. Let $\mathcal{L}_{a}$ be the loss for the audio nodes in the last iGNN block (see Figures \ref{fig:weakly} and \ref{fig:VCL}), then the loss used for EASEE-W is: $\mathcal{L}_{w} = \mathcal{L}_{a}+ \mathcal{L}_{s}+ \mathcal{L}_{c}$. No video labels are required, \ie the speaker-to-speech assignments are unknown.

\subsection{Implementation Details}
We implement  the audio encoder $f_{a}$ with the Resnet18 convolutional encoder \cite{he2016deep} pre-trained on ImageNet \cite{deng2009imagenet}. We adapt the raw 1D audio signal to fit the input of a 2D encoder by generating Mel-frequency cepstral coefficients (MFCCs) of the original audio clip, and then averaging the filters of the network's first convolutional layer to adapt for a single channel input \cite{roth2020ava}. We create the MFCCs with a sampling rate of 16 kHz and an analysis window of 0.025 ms. Our filter bank consists of 26 filters and a fast Fourier transform of size 256 is applied, resulting in 13 cepstrums. The visual encoder $f_{v}$ is based on the R3D architecture,  pre-trained on Kinetics-400 dataset \cite{kay2017kinetics}. For fair comparison with other methods, we also implement $f_{v}$ as a 2D encoder by stacking the temporal and channel dimensions into a single one, then we replicate the filters on the encoder's first layer to accommodate for the input of dimension $(B,CT,H,W)$ \cite{simonyan2014two,roth2020ava}. We also rely on ImageNet pre-training \cite{deng2009imagenet} for this encoder. 

We assemble $\Phi$ on-the-fly with multiple forward passes of $f_{a},f_{v}$, and then map $\Phi$ into nodes of the Graph Convolutional Network and continue with the GCN in a single forward pass. We design the GCN module using the pytorch-geometric library \cite{Fey/Lenssen/2019} and use the EdgeConvolution operator \cite{wang2019dynamic} with filters of size 128. Each GCN layer contains a single iGNN block. EdgeConvolution allows to build a sub-network that performs the message passing between nodes, where every layer (spatial or temporal) in the iGNN is built by a sub-network of two linear layers with ReLu \cite{krizhevsky2012imagenet} and batch normalization \cite{ioffe2015batch}. Therefore, a single iGNN block contains 4 linear layers in total. 

\paragraph{Training EASEE.} We Train EASEE for a total of 12 epochs\footnote{Similar to \cite{alcazar2020active}, we find that sampling every element in the tracklet leads to overfit. For every training epoch, we randomly sample only 4 training examples inside every tracklet.} using the ADAM optimizer \cite{kingma2014adam}, and supervise every node in the final layer with the Cross-Entropy Loss. We also apply intermediate supervision at the end of $f_{a}$ and $f_{v}$ encoders~\cite{roth2020ava}. We empirically observe that this favors faster learning and provides a small performance boost. The learning rate is set to $3 \times 10^{-4}$ and is decreased with annealing $\gamma = 0.1$ at epochs 6 and 8. This very same procedure is applied regardless of the backbone. For every experiment we use a crop size of $160 \times 160$. On average training EASEE-50 requires 28 hours on a single NVIDIA-V100 GPU.

\paragraph{Training EASEE-W.} The training procedure for EASEE-W is similar to that of EASEE. However, we drop all the visual supervision (intermediate supervision of $f_{v}$ included) and reduce the learning rate to $7 \times 10^{-6}$. We apply the assignment loss after the final layer. We calculate this loss individually for every temporal endpoint and accumulate for the full graph. We train for 20 epochs and take learning rate steps at epochs 12 and 16.
\section{Experimental Results}
\begin{table*}[t]
    \centering
    \begin{tabular}{ l l r c c | c }
        \toprule
        & \multicolumn{2}{c}{\textbf{Visual Encoder \quad}} & \textbf{Temporal} &  & \\
        \textbf{Method} & \textbf{Backbone} & \textbf{2D/3D} & \textbf{Context} & \textbf{Supervision} &\textbf{mAP} \\
        \midrule
        EASEE-W (Ours) & Resnet50 & 3D & \cmark & A  & 76.2 \\
        \midrule
        AVA Baseline \etal  \cite{roth2020ava} & MobileNet& 2D & \xmark & A/V  & 79.2 \\
        AVA Baseline + GRU \etal \cite{roth2020ava} & MobileNet & 2D &  \cmark & A/V &82.2 \\
        Chung et al. \cite{chung2019naver} & Custom & 3D+2D & \cmark  & A/V & 85.5 \\
        FaVoA \cite{carneiro2021favoa} & ResNet18 & 2D & \cmark & A/V & 84.7  \\
        MAAS-LAN \cite{leon2021maas} & ResNet18 & 2D & \xmark & A/V & 85.1 \\
        ASC  \cite{alcazar2020active} & Resnet18 & 2D & \cmark & A/V& 87.1  \\
        MAAS-TAN \cite{leon2021maas} & ResNet18 & 2D & \cmark & A/V & 88.8 \\
        EASEE-2D (Ours) & ResNet18 & 2D & \cmark & A/V & 91.1 \\
        UniCon \cite{zhang2021unicon} & Multiple & 2D & \cmark & A/V & 92.0 \\
        
        \midrule
        
        Zhang \etal \cite{zhangmulti} &Custom & 3D+2D & \xmark & A/V & 84.0 \\
        EASEE-50 \textit{\footnotesize l=1, i=3} (\footnotesize Ours) & ResNet50 & 3D & \xmark & A/V & 89.6 \\
        TalkNet \cite{tao2021someone} & Custom & 3D+2D & \cmark & A/V & 92.3 \\
        EASEE-18 (Ours) & ResNet18 & 3D & \cmark & A/V & 93.3 \\
        ASDNet \cite{kopuklu2021design} & ResNext101 & 3D & \cmark & A/V & 93.5 \\
        \textbf{EASEE-50 (Ours)} & ResNet50 & 3D & \cmark & A/V & \textbf{94.1} \\

        \toprule
    \end{tabular}
    \caption{\textbf{State-of-the-art Comparison on AVA-ActiveSpeaker.}  Our best network (EASEE-50) outperforms any other method by at least 0.6 mAP even approaches that build upon much deeper networks. Our smaller network (EASEE-18) remains competitive with the previous state-of-the-art. In the 2D scenario EASEE-2D only lags behind UniCon \cite{zhang2021unicon}, improving the closest method by at least 0.9 mAP. Finally, our weakly supervised configuration (EASEE-W) has a comparable performance with the fully supervised baselines of \cite{roth2020ava}.
    }
    \label{tab:sota}
\end{table*}

In this section, we provide extensive experimental evaluation of our proposed method. We mainly evaluate EASEE on the AVA-ActiveSpeaker dataset~\cite{roth2020ava} and also present additional results on Talkies~\cite{leon2021maas}. We begin with a direct comparison to state-of-the-art methods. Then, we perform an ablation analysis, assessing our main design decisions and their individual contributions to EASEE's final performance. We conclude by presenting the empirical evaluation of EASSE-W in the weakly supervised setup.

The \textbf{AVA-ActiveSpeaker dataset} \cite{roth2020ava} is the first large-scale test-bed for ASD. AVA-ActiveSpeaker contains 262 Hollywood movies: 120 in the training set, 33 in validation, and the remaining 109 in testing. The dataset provides bounding boxes for a total of 5.3 million faces. These face detections are manually linked to produce face tracks that contain a single identity. All AVA-ActiveSpeaker results reported in this paper were obtained using the official evaluation tool provided by the dataset creators, which uses average precision (mAP) as the main metric for evaluation.

\textbf{Talkies} is a manually labeled dataset for the ASD task \cite{leon2021maas}. This dataset was collected from social media videos and contains $23,507$ face tracks extracted from a total of 799,446 individual face detections. Unlike AVA-ActiveSpeaker, it is based on short clips and about 20\% of the speech events are off-screen speech, \ie the event cannot be attributed to a visible person.

\subsection{Comparison to state-of-the-art}\label{subsec:sota}

We compare EASEE against state-of-the-art ASD methods. The results for EASEE are obtained with $l=7$ temporal endpoints, $i=2$ tracklets per endpoint, and a stride of $k=5$. This configuration allows for a sampling window of about 2.41 seconds regardless of the selected backbone. For fair comparison with other methods, we report results of three EASEE variants: `EASEE-50' that uses a 3D backbone based on the ResNet50 architecture, `EASEE-18' that uses a 3D model based on the much smaller Resnet18 architecture, and `EASEE-2D' that uses a 2D Resnet18 backbone. Results are summarized in Table \ref{tab:sota}. 

We find that the optimal number of IGNN blocks changes according to the baseline architecture. For the ResNet18 encoder, 6 blocks (24 layers total in the GCN) are required to achieve the best performance, whereas for ResNet50, only 4 blocks (16 layers total in the GCN) are required. Since we find the best results with $i=2$, and there are scenes with 3 or more simultaneous tracklets, we follow \cite{leon2021maas}. At inference time, we split the speakers in non-overlapping groups of 2, and perform multiple forward passes until every tracklet has been labeled.

We observe that our method outperforms all the other approaches in the validation subset. EASEE-50 is 0.6 mAP higher than the previous state-of-the-art (ASDNet~\cite{kopuklu2021design}). We highlight that ASDNet relies on the deep ResNext101 encoder, whereas EASEE-50 is built on the much smaller ResNet50. Our smaller version (EASEE-18) only lags behind ASDNet by 0.2, and outperforms every other model by at least 1.0 mAP. We also implement a version of EASEE-50 that models only spatial relations (\ie $l=1$). This model reaches 89.6 mAP, outperforming every other network that generates predictions without long-term temporal modeling by at least 4.5 mAP. Finally EASEE-2D outperforms every other 2D approach except UniCon, we explain this result as \cite{zhang2021unicon} presents a far more complex approach that includes multiple 2D backbones to analyze audiovisual data, scene layout and speaker suppression, along with bi-directional GRUs \cite{cho2014properties} for temporal aggregation.

\subsection{Ablation Study}

We ablate our best model (EASEE-50) to assess the individual contributions of our design choices, namely end-to-end training, the iGNN block, and the residual connections between the iGNN blocks.  Table \ref{tab:ablation} contains the individual assessment of each component. The most important architectural design is the end-to-end training, which contributes 1.6 mAP. The proposed iGNN brings about 0.4 mAP when compared against a baseline network where spatial and temporal message passing is performed in the same layer. Finally, residual connections between iGNN blocks contribute with an improved performance of 0.3 mAP. 
\begin{table}[t]
    \centering
    \small
    \begin{tabular}{ l c c c c }
        \toprule
        & &  \textbf{} & \textbf{Residual} & \\
        \textbf{Network}  & \textbf{End-to-End}  & \textbf{iGNN} & \textbf{ Connections}   &\textbf{mAP} \\
        \midrule
        EASEE-50 & \xmark & \xmark & \xmark & 91.9 \\
        EASEE-50 & \cmark & \xmark & \xmark & 93.5 \\
        EASEE-50 & \cmark & \xmark & \cmark & 93.7 \\
        EASEE-50 & \cmark & \cmark & \xmark & 93.8 \\
        EASEE-50 & \cmark & \cmark & \cmark & \textbf{94.1} \\
        \toprule
    \end{tabular}
    \caption{\textbf{AVA-ActiveSpeaker Ablation.} We assess the empirical contribution of the most relevant components in EASEE. Residual connections contribute about 0.3 mAP and the proposed iGNN block 0.4 mAP. Overall the most relevant design choice is the end-to-end trainable nature of EASEE contributing 1.6 mAP.
    }
    \label{tab:ablation}
\end{table}

\paragraph{Intermediate Embedding Configuration.}
We compare the performance of EASEE-50 with different  configurations of the intermediate embedding $\Phi$. In Table \ref{tab:speakers_1}, we assess the performance of EASEE-50 when changing the number of temporal endpoints $l$ and the number of tracklets $i$.

We observe that the best performance arises when $i=2$, which is in stark contrast to other methods \cite{kopuklu2021design,leon2021maas,zhang2021unicon} that often rely on aggregating information from 4 or more visual tracklets. We attribute this to the end-to-end nature of EASEE, where contextual cues are directly optimized for the ASD problem, thus requiring less spatial data for effective predictions. Nonetheless, for small values of $l$, we find that EASEE actually benefits from a larger number of visual tracklets ($i=3$). This suggests that in the absence of strong temporal cues, EASEE will focus on extracting meaningful information from the spatially adjacent tracklets.

We also observe that the temporal dimension of the problem (number of endpoints $l$) is more relevant than the spatial component (number of concurrent tracklets $i$). When increasing $l$ from 1 to 7, performance improves significantly, by 4.8 mAP on average. In contrast, increasing visual tracklets from $i=1$ to $i=4$ only yields 1.1 mAP improvement on average. This is consistent with related works, which show a performance boost when incorporating recurrent Units and long temporal samplings~\cite{roth2020ava,alcazar2020active,kopuklu2021design}. 

In Table \ref{tab:clip_1}, we analyze the effect of the input clip size to the encoder $f_{v}$ in EASEE. We find that as the clip size increases, performance also improves but saturates around 15 frames (about 0.62 seconds). For every clip size, longer temporal sampling (more endpoints) provides better results. The best result is achieved at $l=7$ with clips of 15 frames. 

\begin{table}[t]
    \centering
    \begin{tabular}{ r | c c c c c}
        \toprule
        & \multicolumn{5}{c}{ \textbf{End Points ($l$)} } \\
        \textbf{Speakers ($i$)} & $1$ \quad & $3$ \quad& $5$ \quad& $7$ & $9$ \quad\\ 
        \midrule
        $1$     \quad  & $86.6$ & $90.5$ & $92.4$ & $92.9$ & $92.6$    \\
        $2$     \quad  & $89.0$ & $92.3$ & $93.4$ & \textbf{94.1} & $93.8$   \\ 
        $3$     \quad  & $89.6$ & $92.2$ & $93.3$ & $93.8$ & $93.4$   \\ 
        $4$     \quad  & $89.2$ & $91.8$ & $93.1$ & $93.7$ & $93.2$    \\ 
        \bottomrule
    \end{tabular}
    \caption{
        \textbf{End-Points vs speaker.} In EASEE longer temporal windows allow to improve the performance, achieving the best result at $l=7$. A large number of speakers favors performance in shorter windows but $i=2$ is the best parameter for long windows ($l \geq 3$).
    }
    \label{tab:speakers_1}
\end{table}

\begin{table}[t]
    \centering
    \begin{tabular}{ r | c c c c c}
        \toprule
        & \multicolumn{5}{c}{ \textbf{End Points ($l$)} } \\
        \textbf{Clip Size} & $1$ \quad & $3$ \quad & $5$ \quad & $7$ \quad& $9$ \quad\\ 
        \midrule
    
        $11$  \quad  & $87.6$ & $91.4$ & $93.1$ & $93.5$ & $93.2$\\ 
        $13$  \quad  & $88.3$ & $91.7$ & $93.3$ & $93.9$ & $93.6$  \\ 
        $15$  \quad  & $89.0$ & $92.3$ & $93.4$ & \textbf{94.1} & $93.8$   \\ 
        $17$  \quad  & $89.3$ & $92.5$ & $93.3$ & $93.9$ & $93.7$ \\ 
        \bottomrule
    \end{tabular}
    \caption{ 
        \textbf{End-Points vs Input Clip.} Long temporal sampling enables better predictions in most scenarios. In the EASEE architecture the input size for the 3D encoder also provides improved performance, the optimal is 15 frames, which equals 0.62 seconds.  
    }
    \label{tab:clip_1}
\end{table}

\paragraph{Design of iGNN blocks.}

We begin the empirical assessment of the iGNN module by analyzing two architectural decisions in EASEE: i) The effect of the number of iGNN modules, and ii) the size (number of neurons) in the linear layers in the iGNN blocks.

We first analyze the effect of the number of iGNN blocks. We control this hyper-parameter for the Resnet50 Backbone and the Resnet18 Backbone, and evaluate from 2 to 7 iGNN modules. Table \ref{tab:iGNNABlat} summarizes the results. Deeper GNN networks lead to higher performance, but this improvement stalls at 4 iGNN blocks for the Resnet50 backbone and 6 iGNN blocks for the Resnet18.

We conclude by analyzing the effect of the size of the linear layers used in iGNN. Our best models (EASEE-50 \& EASEE-18) use linear layers of size 128. In table \ref{tab:filters} we ablate the size of this layer in the EASE50 architecture. We see a smaller impact on this hyper-parameter, where a smaller net only loses $0.3$ mAP, and iGNN blocks with double the number of neurons only lose $0.2$ mAP.

\begin{table*}[t]
    \centering
    \begin{tabular}{c c c c c c c}
        \hline
        \textbf{Backbone} & \textbf{2 iGNN} & \textbf{3 iGNN}  & \textbf{4 iGNN}  &  \textbf{5 iGNN}  & \textbf{6 iGNN} & \textbf{7 iGNN}  \\
        \hline 
        EASEE-18 & 92.8 & 93.0 & 93.2 & 93.2 & \textbf{93.3} & 93.2\\
        EASEE-50 & 93.6 & 93.8& \textbf{94.1} & 94.0 & 93.8 & 93.8  \\
        \hline
    \end{tabular}
    \caption{
        \textbf{EASEE Performance By iGNN Depth.} We analyze the effect of the number of iGNN blocks in EASEE. Stacking blocks improves the performance util 4 blocks are stacked (Resnet50) or 6 blocks are stacked (Resnet18)
    }
    \label{tab:iGNNABlat}
\end{table*}

\begin{table}[t]
    \centering
    \begin{tabular}{c c c c c}
        \hline
        \textbf{Layer Size} \quad & \textbf{64} \quad & \textbf{128} \quad & \textbf{224} \quad  & \textbf{256}  \\
        \hline 
        EASEE-50 & 93.8 & \textbf{94.1} & 93.9 & 93.9   \\
        \hline
    \end{tabular}
    \caption{
        \textbf{Linear layer size.} We assess the effect of the layer size in the iGNN module. We find slightly reduced performance by altering the size of the iGNN module. 
    }
    \label{tab:filters}
\end{table}

\begin{table}[t]
    \centering
    \footnotesize
    \begin{tabular}{c c c c c}
    	\toprule
        \textbf{iGNN} & \textbf{iGNN-TS}  & \textbf{Two Stream} & \textbf{Parallel}  & \textbf{Spatio-Temp.} \cite{leon2021maas} \\
        \midrule
        94.1 & 94.0 & 92.8 & 93.7 & 93.7 \\
    	\toprule
    \end{tabular}
    \caption{
        \textbf{iGNN Layering Strategies.} We compare multiple strategies to assemble our iGNN block, we find that interleaving the temporal and spatial messages brings the best results. In comparison a joint massage passing will reduce the performance by 0.4 mAP, a naive join of these steps with a linear layer reports the same performance reduction.
    }
    \label{tab:ignn_style}
\end{table}

After testing the size configurations of the iGNN block, we assess the effectiveness of the proposed iGNN block by comparing it against the following fusion alternatives: (a) Temporal-Spatial (iGNN-TS), an immediate alternative to iGNN where temporal message passing is performed before any spatial message passing is done; (b) Two Stream, where two independent GCN streams  perform spatial and  temporal message passing respectively, and  these streams are fused at the end of the network; (c) Parallel, where the block performs spatial and temporal message passing in parallel and fuses the features using a fully connected layer; (d) Spatio-Temporal, where a single graph structure performs temporal and spatial message passing at the same time \cite{leon2021maas}. Table \ref{tab:ignn_style} summarizes the results.

Overall, we find that the best block design is the one in which spatial message passing occurs first. Reversing the order of message passing results in a very similar alternative with only minor performance degradation. In comparison, the two-stream approach performs significantly worse than all other alternatives, suggesting that the fusion of temporal and spatial information must occur earlier to be effective in an end-to-end scenario. Joint spatio-temporal messaging also has high performance, but still lags behind iGNN.

\vspace{-0.25cm}
\paragraph{Talkies Dataset} We conclude this subsection with the evaluation of EASEE on the Talkies dataset \cite{leon2021maas}. Here, we test: (i) a direct transfer of EASEE-50 into the validation set of Talkies, (ii) directly training EASEE on Talkies, and (iii) using Talkies as downstream task after pre-training on AVA-ActiveSpeaker. Table \ref{tab:talkies} summarizes the results.

EASEE outperforms \cite{leon2021maas} for the direct transfer on the Talkies dataset. Moreover,  training on Talkies results in a high performance comparable to that of the AVA-ActiveSpeaker dataset, this is particularly interesting as Talkies is a dataset that contains a large portion of scenes with out-of screen speech, a situation that is extremely rare in the AVA-Active Speaker. Finally, using talkies as a downstream task results in 1.0 mAP improvement, which is about 15\% relative error improvement.

\section{Weak Supervision}
\label{subsec:weaksupervision}
We conclude this section by evaluating the weakly supervised version of EASEE, \ie EASEE-W. To the best of our knowledge, there are no comparable methods that strictly rely on weak (audio only) supervision in the ASD task. Therefore, we establish multiple baselines, from random predictions to direct speech-to-speaker assignment. 

We first consider baselines that ignore audio labels and the structure of the ASD problem: i) \textit{random} baseline where every speaker gets a random score sampled from a uniform distribution between $[0,1]$.  ii) \textit{Naive Recall} where we trivially predict every tracklet as an active speaker and iii) \textit{Naive Precision} that trivially predicts every tracklet as silent. We also build baselines that rely on audio supervision. We use our trained audio encoder $f_{a}$ to detect speech intervals and generate random speech-to-speaker assignments within that time window. We explore two approaches: iv) \textit{Naive Audio assignment} where we choose a random visible speaker whenever a speech event is detected. v) \textit{Largest Face Audio assignment} since AVA-Active Speaker is a collection of Hollywood movies, we follow a common bias in commercial movies, and assign the speech event to the tracklet that occupies the largest area in the screen. Table \ref{tab:weak} summarizes the results of these experiments in the AVA-ActiveSpeaker dataset.

\begin{table}[t]
    \centering

    \begin{tabular}{ l c c c c }
        \toprule
        \textbf{} & ~~\textbf{AVA} & ~~\textbf{Talkies } & ~~\textbf{} \\
        \textbf{Network} & ~~\textbf{Pre-train} & ~~\textbf{Training} & ~~\textbf{mAP} \\
        \midrule
        MAAS-TAN \cite{leon2021maas} & \cmark & \xmark & 79.1 \\
        EASEE-50 & \cmark & \xmark & 86.7 \\
        EASEE-50 & \xmark & \cmark & 93.6 \\
        EASEE-50 & \cmark & \cmark & 94.5 \\
       
        \toprule
    \end{tabular}
    \caption{\textbf{Evaluation on Talkies Dataset.} We evaluate EASEE on the Talkies dataset. It outperforms the existing baseline on the direct transfer from AVA-ActiveSpeaker, and show the results of training EASEE end-to-end in Talkies. Finally we test the effectiveness of AVA-ActiveSpeakers as pre-training for Talkies.
    }
    \label{tab:talkies}
\end{table}
\begin{table}[t]
    \centering
    \begin{tabular}{ l c c c c }
        \toprule
        \textbf{Network} &\textbf{mAP} \\
        \midrule
        Random & 25.1\\
        Naive Recall & 27.1\\
        Naive Precision & 27.1\\
        \midrule
        Naive Audio Assignment &  47.7\\
        Large Face Audio Assignment & 49.1\\
        \midrule
        EASEE-W $L_{a}$ only & 26.1\\
        EASEE-W $L_{a}, L_{c}$ only & 25.8 \\
        EASEE-W $L_{a}, L_{s}$ only & 54.4 \\
        EASEE-W  ($L_{a}, L_{s}, L_{c}$) & 76.2\\
        Fully Supervised 2D encoder \cite{roth2020ava,alcazar2020active} & 79.5\\
        \toprule
    \end{tabular}
    \caption{\textbf{Weak Supervision.} We show that EASEE-W largely improves over baseline approaches for ASD, it outperforms a naive baseline by 28.5 mAP, and remains competitive with fully supervised 2D encoder.  
    }
    \label{tab:weak}
\end{table}

Overall, we see that random baselines largely under-perform. Even when the predictions have a bias towards the largest class (silent) results are just 27.1 mAP. A relevant increment in performance (about 20 mAP) appears when the audio supervision is used to generate the naive visual assignments. This improvement is a direct result of the structure in the ASD problem, where speech events are attributed to a defined set of sources.

When we apply EASEE-W, we see the complementary behavior of the proposed loss functions. The baseline with audio supervision ($L_a$ only) exhibits no meaningful improvement over the random base, despite the GCN structure.  A similar situation can be observed if we use the audio supervision and enforce temporal consistency on the visual features ($L_{c}$). This indicates that information flow across modalities can not be trivially enforced by the GCN module or temporal visual consistency. Including the assignment loss ($L_s$) results in a scenario that already improves over the naive assignments suggesting that local attributions already favors some meaningful audiovisual patterns. Finally, the best result is achieved when assignments and temporal consistency for the visual data are considered. This result improves over any baseline by at least 27 mAP. We conclude this section highlighting that this result is competitive with baseline approaches that rely on encoding short-temporal information from a single speaker as outlined in \cite{roth2020ava,alcazar2020active}.

\paragraph{EASEE-W Ablation Analysis} We conclude this section by ablating EASEE-W according to the face size and number of simultaneous tracklets in the AVA-ActiveSpeaker validation set. We compare EASEE-W performance against the baseline approach of \cite{roth2020ava}. Despite the large difference in the training and supervision signals, EASEE-W follows a similar error pattern to the fully supervised method of \cite{roth2020ava}.

\begin{table}[h]
    \scriptsize
    \centering
    \begin{tabular}{l | c c c c c c c}
        \hline
        \textbf{Network} & \textbf{Large} & \textbf{Med.} & \textbf{Small} & \textbf{1 Face} & \textbf{2 Faces} & \textbf{3 Faces} \\
        \hline
        EASEE-W & 87.3& 72.6 & 42.2  & 87.8 & 64.5 &  52.6  \\
        Roth et al. \cite{roth2020ava} & 92.2 & 79.0 & 56.2 & 87.9 & 71.6 & 54.4 \\
    	\toprule
    \end{tabular}
\end{table}

\section{Additional Experimental Results}

We complement the analysis of EASEE, and assess its performance in known challenging scenarios. We follow the procedure of \cite{roth2020ava}, and evaluate EASEE in the AVA-ActiveSpeakers dataset according to: i) number of visible faces, and ii) the size of the face.

Table \ref{tab:ava_size}  shows the ablation of the  performance of EASEE according to the face size. Overall, EASEE shows a similar behavior to state-of-the-art methods, where smaller faces (less than $64 \times 64$) are harder to classify (79.3 mAP). Medium images  (between $64 \times 64$ and $128 \times 128$) show an improvement in performance over small images, and large faces report the highest mAP at {97.7 mAP}.

\begin{table}[h]
    \centering
    \footnotesize
      \begin{tabular}{l c c c c }
        \hline
        \textbf{Faces Size} \qquad & \textbf{EASEE-50} & \textbf{ASDNet} \cite{kopuklu2021design} & \textbf{MAAS} \cite{leon2021maas} & \textbf{ASC}  \cite{alcazar2020active} \\
        \hline 
        Small & \textbf{79.3} & 74.3 & 55.2 & 44.9 \\
        Medium & \textbf{93.2} & 89.8 & 79.4 & 68.3  \\
        Large &  \textbf{97.7} & 96.3 & 93.0 & 86.4  \\
        \hline
    \end{tabular}
    \caption{
        \textbf{AVA-ActiveSpeaker Face Size.}
        We evaluate EASEE in the AVA-ActiveSpeaker dataset according to the size of the faces. As observed in previous works smaller faces are harder to classify. EASEE
        outperforms the state-of-the art in every scenario
    }
    \label{tab:ava_size}
\end{table}

Table \ref{tab:avanumsp} evaluates the performance of EASEE according to the number of simultaneous faces. Just like other ensemble methods, EASEE shows an improved performance in the multi-speaker scenario when compared to the single speaker baseline \cite{roth2020ava} (20.8 mAP improvement for two speakers, 29.5 mAP improvement for 3 speakers). 

\begin{table}[h]
    \centering
    \footnotesize
    \begin{tabular}{c c c c c}
        \hline
        \textbf{Number} & & & &  \\
        \textbf{of Faces} \quad & \textbf{EASEE-50} & \textbf{ASDNet} \cite{kopuklu2021design} & \textbf{MAAS} \cite{leon2021maas}  &  \textbf{ASC} \cite{alcazar2020active} \\
        \hline 
        1 &  \textbf{96.5} & 95.7 &93.3 & 91.8 \\
        2 &  \textbf{92.4} & 92.4 & 85.8 & 83.8   \\
        3 &  \textbf{83.9} & 83.7 & 68.2 & 67.6  \\
        \hline
    \end{tabular}
    \caption{
        \textbf{Performance evaluation by number of faces.} 
        We evaluate EASEE in the AVA-ActiveSpeaker according to the number of visible faces (tracklets) in the scene. Multi-speaker scenes are far more challenging, our method outperforms the current state-of-the-art in any scenario.
    }
    \label{tab:avanumsp}
\end{table}

\section{Conclusion}
We introduced EASEE, a multi-modal end-to-end trainable network for the ASD task. EASEE outperforms state-of-the-art approaches in the large scale AVA-ActiveSpeaker\cite{roth2020ava} dataset, and transfers effectively to smaller sets that contain out-of-screen speech. EASEE allows for fully supervised and weakly supervised training by leveraging the inherent structure of the ASD problem and the natural consistency in video data. Future explorations on the ASD problem might rely on our label efficient training setup.

{\small
\bibliographystyle{ieee_fullname}
\bibliography{EASEE}
}

\end{document}